\documentclass[10pt,twocolumn,letterpaper,table,dvipsnames]{article}
\usepackage[table, dvipsnames]{xcolor}
\usepackage{iccv}
\usepackage{times}
\usepackage{epsfig}
\usepackage{graphicx}
\usepackage{amsmath}
\usepackage{amssymb}
\usepackage{enumitem}
\usepackage[accsupp]{axessibility}

\usepackage{booktabs}
\usepackage{multirow}
\usepackage{xspace}
\usepackage{flushend}  % keeps references balanced on last page

\newcommand{\best}[1]{\textcolor{Red}{\textbf{#1}}}
\newcommand{\second}[1]{\textcolor{Orange}{#1}}
\newcommand{\methodname}{SAFE\xspace}

\usepackage[pagebackref=true,breaklinks=true,colorlinks,bookmarks=false]{hyperref}

\iccvfinalcopy % *** Uncomment this line for the final submission

\def\iccvPaperID{5333} % *** Enter the ICCV Paper ID here

\ificcvfinal\fi

\AddToShipoutPicture*{%
     \AtTextUpperLeft{%
         \put(-3.5,10){
           \begin{minipage}{\textwidth}
              \scriptsize
              \MakeUppercase{Preprint version of an IEEE International Conference on Computer Vision (ICCV) article\\final version available at} \url{https://ieeexplore.ieee.org}
           \end{minipage}}%
     }%
}

\begin{document}

%%%%%%%%% TITLE
\title{SAFE: Sensitivity-Aware Features for \\Out-of-Distribution Object Detection}

\author{Samuel Wilson$^1$, Tobias Fischer$^1$, Feras Dayoub$^2$, Dimity Miller$^1$, Niko S\"underhauf$^1$\\
$^1$QUT Centre for Robotics, Queensland University of Technology \\
$^2$Australian Institute for Machine Learning, University of Adelaide \\
{\tt\small s84.wilson@hdr.qut.edu.au}
% For a paper whose authors are all at the same institution,
% omit the following lines up until the closing ``}''.
% Additional authors and addresses can be added with ``\and'',
% just like the second author.
% To save space, use either the email address or home page, not both
% \and
% Tobias Fischer\\
% Queensland University of Technology\\
% 2 George St, Brisbane, QLD 4000, Australia\\
% {\tt\small tobias.fischer@qut.edu.au}
% \and
% Feras Dayoub\\
% University of Adelaide\\
% North Terrace, Adelaide, SA 5005, Australia\\
% {\tt\small feras.dayoub@adelaide.edu.au}
% \and
% Dimity Miller\\
% Queensland University of Technology\\
% 2 George St, Brisbane, QLD 4000, Australia\\
% {\tt\small d24.miller@qut.edu.au}
% \and
% Niko Sünderhauf\\
% Queensland University of Technology\\
% 2 George St, Brisbane, QLD 4000, Australia\\
% {\tt\small niko.suenderhauf@qut.edu.au}
}

\maketitle
% Remove page # from the first page of camera-ready.
\ificcvfinal\thispagestyle{empty}\fi

\newcommand{\vect}[1]{\mathbf{ #1}}
\newcommand{\vectg}[1]{{\boldsymbol{ #1}}}
\newcommand{\ggo}{\ensuremath{\mathrm{g^2o}} }
\newcommand{\R}{\mathbb{R}}
\newcommand{\I}{\mathbb{I}}
\newcommand{\N}{\mathbb{N}}
\newcommand{\Z}{\mathbb{Z}}
\renewcommand{\P}{\mathbb{P}}
\newcommand{\tran}{^\top}
\newcommand{\T}{^\mathsf{T}}
\newcommand{\iT}{^{-\mathsf{T}}}
\newcommand{\inv}{^{-1}}
\newcommand{\func}[2]{\mathtt{#1}\left\{#2\right\}}
\newcommand{\sig}{\operatorname{sig}}
\newcommand{\diag}{\operatorname{diag}}
\newcommand{\argmin}{\operatornamewithlimits{argmin}}
\newcommand{\argmax}{\operatornamewithlimits{argmax}}
\newcommand{\RMSE}{\operatorname{RMSE}}
\newcommand{\RMSEpos}{\operatorname{RMSE}_\text{pos}}
\newcommand{\RMSEori}{\operatorname{RMSE}_\text{ori}}
\newcommand{\RPE}{\operatorname{RPE}}
\newcommand{\RPEpos}{\operatorname{RPE}_\text{pos}}
\newcommand{\RPEori}{\operatorname{RPE}_\text{ori}}
\newcommand{\rpe}{\varepsilon_{\vdelta}}
\newcommand{\achiError}{\bar{e}_{\chi^2}}
\newcommand{\chiError}{e_{\chi^2}}
\newcommand{\normal}[2]{\mathcal{N}\left(#1, #2\right)}
\newcommand{\uniform}[2]{\mathcal{U}\left(#1, #2\right)}
\newcommand{\pfrac}[2]{\frac{\partial #1}{\partial #2}}  %
\newcommand{\fracpd}[2]{\frac{\partial #1}{\partial #2}} %
\newcommand{\fracppd}[2]{\frac{\partial^2 #1}{\partial #2^2}}  %
\newcommand{\dd}{\mathrm{d}}  
\newcommand{\smd}[2]{\left\| #1 \right\|^2_{#2}}
\newcommand{\E}[1]{\text{\normalfont{E}}\left[ #1 \right]}     %
\newcommand{\Cov}[1]{\text{\normalfont{Cov}}\left[ #1 \right]} %
\newcommand{\Var}[1]{\text{\normalfont{Var}}\left[ #1 \right]} %
\newcommand{\Tr}[1]{\text{\normalfont{tr}}\left( #1 \right)}   %
\def\sgn{\mathop{\mathrm sgn}}    
\newcommand{\twovector}[2]{\begin{pmatrix} #1 \\ #2 \end{pmatrix}} %
\newcommand{\smalltwovector}[2]{\left(\begin{smallmatrix} #1 \\ #2 \end{smallmatrix}\right)} 
\newcommand{\threevector}[3]{\begin{pmatrix} #1 \\ #2 \\ #3 \end{pmatrix}} %
\newcommand{\fourvector}[4]{\begin{pmatrix} #1 \\ #2 \\ #3 \\ #4 \end{pmatrix}}  %
\newcommand{\smallthreevector}[3]{\left(\begin{smallmatrix} #1 \\ #2 \\ #3 \end{smallmatrix}\right)} %
\newcommand{\fourmatrix}[4]{\begin{pmatrix} #1 & #2 \\ #3 & #4 \end{pmatrix}} %
\newcommand{\vA}{\vect{A}}
\newcommand{\vB}{\vect{B}}
\newcommand{\vC}{\vect{C}}
\newcommand{\vD}{\vect{D}}
\newcommand{\vE}{\vect{E}}
\newcommand{\vF}{\vect{F}}
\newcommand{\vG}{\vect{G}}
\newcommand{\vH}{\vect{H}}
\newcommand{\vI}{\vect{I}}
\newcommand{\vJ}{\vect{J}}
\newcommand{\vK}{\vect{K}}
\newcommand{\vL}{\vect{L}}
\newcommand{\vM}{\vect{M}}
\newcommand{\vN}{\vect{N}}
\newcommand{\vO}{\vect{O}}
\newcommand{\vP}{\vect{P}}
\newcommand{\vQ}{\vect{Q}}
\newcommand{\vR}{\vect{R}}
\newcommand{\vS}{\vect{S}}
\newcommand{\vT}{\vect{T}}
\newcommand{\vU}{\vect{U}}
\newcommand{\vV}{\vect{V}}
\newcommand{\vW}{\vect{W}}
\newcommand{\vX}{\vect{X}}
\newcommand{\vY}{\vect{Y}}
\newcommand{\vZ}{\vect{Z}}
\newcommand{\va}{\vect{a}}
\newcommand{\vb}{\vect{b}}
\newcommand{\vc}{\vect{c}}
\newcommand{\vd}{\vect{d}}
\newcommand{\ve}{\vect{e}}
\newcommand{\vf}{\vect{f}}
\newcommand{\vg}{\vect{g}}
\newcommand{\vh}{\vect{h}}
\newcommand{\vi}{\vect{i}}
\newcommand{\vj}{\vect{j}}
\newcommand{\vk}{\vect{k}}
\newcommand{\vl}{\vect{l}}
\newcommand{\vm}{\vect{m}}
\newcommand{\vn}{\vect{n}}
\newcommand{\vo}{\vect{o}}
\newcommand{\vp}{\vect{p}}
\newcommand{\vq}{\vect{q}}
\newcommand{\vr}{\vect{r}}
\newcommand{\vt}{\vect{t}}
\newcommand{\vu}{\vect{u}}
\newcommand{\vv}{\vect{v}}
\newcommand{\vw}{\vect{w}}
\newcommand{\vx}{\vect{x}}
\newcommand{\vy}{\vect{y}}
\newcommand{\vz}{\vect{z}}
\newcommand{\valpha}{\vectg{\alpha}}
\newcommand{\vbeta}{\vectg{\beta}}
\newcommand{\vgamma}{\vectg{\gamma}}
\newcommand{\vdelta}{\vectg{\delta}}
\newcommand{\vepsilon}{\vectg{\epsilon}}
\newcommand{\vtau}{\vectg{\tau}}
\newcommand{\vmu}{\vectg{\mu}}
\newcommand{\vphi}{\vectg{\phi}}
\newcommand{\vPhi}{\vectg{\Phi}}
\newcommand{\vpi}{\vectg{\pi}}
\newcommand{\vPi}{\vectg{\Pi}}
\newcommand{\vPsi}{\vectg{\Psi}}
\newcommand{\vchi}{\vectg{\chi}}
\newcommand{\vvarphi}{\vectg{\varphi}}
\newcommand{\veta}{\vectg{\eta}}
\newcommand{\viota}{\vectg{\iota}}
\newcommand{\vkappa}{\vectg{\kappa}}
\newcommand{\vlambda}{\vectg{\lambda}}
\newcommand{\vLambda}{\vectg{\Lambda}}
\newcommand{\vnu}{\vectg{\nu}}
\newcommand{\vgo}{\vectg{\o}}
\newcommand{\vvarpi}{\vectg{\varpi}}
\newcommand{\vtheta}{\vectg{\theta}}
\newcommand{\vTheta}{\vectg{\Theta}}
\newcommand{\vvartheta}{\vectg{\vartheta}}
\newcommand{\vrho}{\vectg{\rho}}
\newcommand{\vsigma}{\vectg{\sigma}}
\newcommand{\vSigma}{\vectg{\Sigma}}
\newcommand{\vvarsigma}{\vectg{\varsigma}}
\newcommand{\vupsilon}{\vectg{\upsilon}}
\newcommand{\vomega}{\vectg{\omega}}
\newcommand{\vOmega}{\vectg{\Omega}}
\newcommand{\vxi}{\vectg{\xi}}
\newcommand{\vXi}{\vectg{\Xi}}
\newcommand{\vpsi}{\vectg{\psi}}
\newcommand{\vzeta}{\vectg{\zeta}}
\newcommand{\vzero}{\vect{0}}
\newcommand{\cA}{\mathcal{A}}
\newcommand{\cB}{\mathcal{B}}
\newcommand{\cC}{\mathcal{C}}
\newcommand{\cD}{\mathcal{D}}
\newcommand{\cE}{\mathcal{E}}
\newcommand{\cF}{\mathcal{F}}
\newcommand{\cG}{\mathcal{G}}
\newcommand{\cH}{\mathcal{H}}
\newcommand{\cI}{\mathcal{I}}
\newcommand{\cJ}{\mathcal{J}}
\newcommand{\cK}{\mathcal{K}}
\newcommand{\cL}{\mathcal{L}}
\newcommand{\cM}{\mathcal{M}}
\newcommand{\cN}{\mathcal{N}}
\newcommand{\cO}{\mathcal{O}}
\newcommand{\cP}{\mathcal{P}}
\newcommand{\cQ}{\mathcal{Q}}
\newcommand{\cR}{\mathcal{R}}
\newcommand{\cS}{\mathcal{S}}
\newcommand{\cT}{\mathcal{T}}
\newcommand{\cU}{\mathcal{U}}
\newcommand{\cV}{\mathcal{V}}
\newcommand{\cW}{\mathcal{W}}
\newcommand{\cX}{\mathcal{X}}
\newcommand{\cY}{\mathcal{Y}}
\newcommand{\cZ}{\mathcal{Z}}
\newcommand{\fA}{\mathfrak{A}}
\newcommand{\fB}{\mathfrak{B}}
\newcommand{\fC}{\mathfrak{C}}
\newcommand{\fD}{\mathfrak{D}}
\newcommand{\fE}{\mathfrak{E}}
\newcommand{\fF}{\mathfrak{F}}
\newcommand{\fG}{\mathfrak{G}}
\newcommand{\fH}{\mathfrak{H}}
\newcommand{\fI}{\mathfrak{I}}
\newcommand{\fJ}{\mathfrak{J}}
\newcommand{\fK}{\mathfrak{K}}
\newcommand{\fL}{\mathfrak{L}}
\newcommand{\fM}{\mathfrak{M}}
\newcommand{\fN}{\mathfrak{N}}
\newcommand{\fO}{\mathfrak{O}}
\newcommand{\fP}{\mathfrak{P}}
\newcommand{\fQ}{\mathfrak{Q}}
\newcommand{\fR}{\mathfrak{R}}
\newcommand{\fS}{\mathfrak{S}}
\newcommand{\fT}{\mathfrak{T}}
\newcommand{\fU}{\mathfrak{U}}
\newcommand{\fV}{\mathfrak{V}}
\newcommand{\fW}{\mathfrak{W}}
\newcommand{\fX}{\mathfrak{X}}
\newcommand{\fY}{\mathfrak{Y}}
\newcommand{\fZ}{\mathfrak{Z}}

\begin{abstract}

We address the problem of out-of-distribution (OOD) detection for the task of object detection. We show that residual convolutional layers with batch normalisation produce \textbf{S}ensitivity-\textbf{A}ware \textbf{FE}atures (\methodname) that are consistently powerful for distinguishing in-distribution from out-of-distribution detections. We extract SAFE vectors for every detected object, and train a multilayer perceptron on the surrogate task of distinguishing adversarially perturbed from clean in-distribution examples. This circumvents the need for realistic OOD training data, computationally expensive generative models, or retraining of the base object detector. \methodname outperforms the state-of-the-art OOD object detectors on multiple benchmarks by large margins, e.g.~reducing the FPR95 by an absolute 30.6\% from 48.3\% to 17.7\% on the OpenImages dataset.
\end{abstract}

\section{Introduction}
Across a variety of tasks, deep neural networks (DNNs) produce state-of-the-art performance when tested on data that closely matches the training data distribution \cite{He_2016_CVPR, simonyan2015a}. However, when deployed into the real world, out-of-distri\-bution (OOD) samples that do not belong to the training distribution are likely encountered. Upon encountering OOD samples, DNNs tend to fail silently and produce overconfident erroneous predictions~\cite{DBLP:journals/corr/abs-1808-03305, miller2018dropout, hendrycks17baseline, bendale2016towards, 9093355, pmlr-v70-guo17a}. Especially in safety-critical applications, such as self-driving vehicles or medical robotics, such silent failures present a severe safety risk that must be addressed before the widespread adoption of these systems~\cite{sunderhauf2018limits, amodei2016concrete}. 

OOD detection, where OOD samples are distinguished from in-distribution (ID) samples, is thus an important task. OOD detection has been addressed widely in the image \emph{classification} setting~\cite{Mahalanobis, liu2020energy, OutlierExposure, pmlr-v119-sastry20a, HDFF, NMD, sun2021react, yang2022openood}.%
\begin{figure}
    \centering
    \includegraphics[width=\columnwidth]{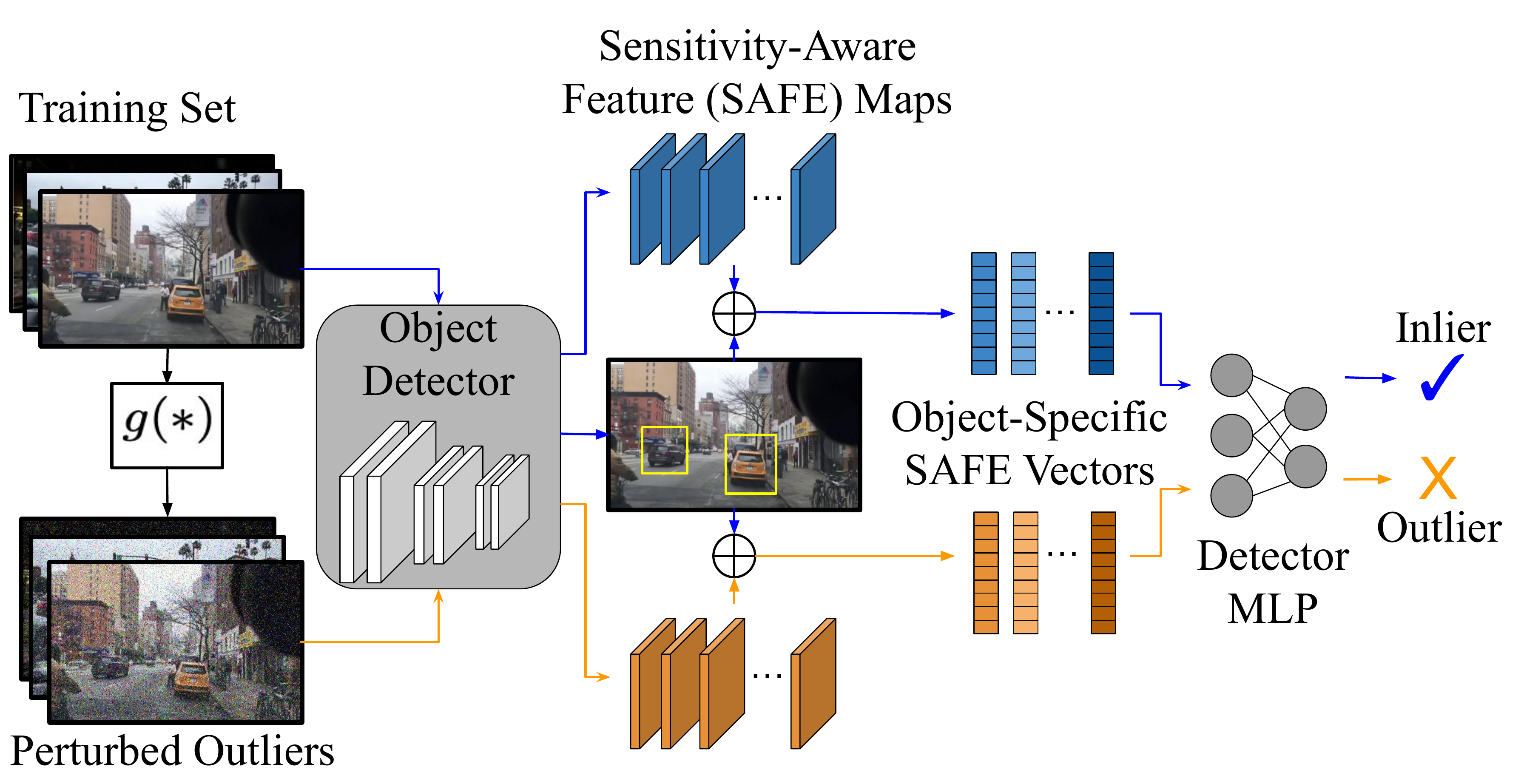}
    \caption{
    Overview of our proposed \methodname OOD object detector. Feature maps are extracted from sensitivity-aware layers in the backbone of a pretrained object detector. Object-specific \methodname vectors are extracted for the predicted bounding boxes. Pre-deployment, an auxiliary MLP is trained to distinguish the feature vectors of normal ID detections (\textcolor{Blue}{blue}) from adversarially-perturbed ID samples (\textcolor{Orange}{orange}). At test time, the pipeline for the training samples is repeated (\textcolor{Blue}{blue}) for all test samples, with the auxiliary MLP producing OOD detection scores for each object in a test image. Illustrative input images are drawn from BDD100K~\cite{BDD100K}.
    }
    \label{fig:hook}
\end{figure}
In this paper, we expand upon the limited body of work in OOD \emph{object detection}~\cite{vos, du2022stud, du2022siren}, leveraging recent theoretical insights on the behaviour of the feature space of DNNs. Specifically, recent theory~\cite{SNGP, DDU, DUQ, ResnetInputDistance} has highlighted the importance of ensuring that the feature space of a DNN is \emph{distance-preserving} through the constraints of \emph{sensitivity} and \emph{smoothness}. In particular, \emph{sensitivity}, \ie the preservation of input distances in the output, has been shown to play a crucial role in learning a robust feature set that avoids mapping ID and OOD data to similar feature representations~\cite{DUQ}. 

Furthermore, prior work established the role of adversarial attacks in OOD generalisation~\cite{generalOOD}, increasing the separability of ID samples from OOD~\cite{ODIN, General-ODIN} and perturbing feature representations~\cite{FGSM, advAttacksSurvey, advmedical}. We thus leverage the most \emph{sensitive} layers in a pretrained object detector backbone through targeted input-level adversarial perturbations. %

This paper introduces \textbf{\methodname}, a new approach to visual OOD object detection that utilises object-specific \textbf{S}ensitivity-\textbf{A}ware \textbf{Fe}atures (Figure~\ref{fig:hook}). Our approach has three core components, each offering advantages over existing works in this area:
\begin{enumerate}[nosep] 
    \item We identify a critical subset of layers that are \emph{sensitive} to OOD input variations, \ie these layers preserve differences from the input in their feature space, and trigger abnormally high activations. \methodname layers are residual convolutional layers with batch normalisation within the backbone of an object detector. We empirically validate their superiority for OOD detection in our results. In contrast, previous work only uses features from the classification head of the object detector and do not consider the characteristics of different layers for OOD detection~\cite{vos,pmlr-v119-sastry20a,Mahalanobis}. 
    \item We extract object-specific \methodname vectors and use a multi-layer perceptron (MLP) to classify every detected object as ID or OOD. This allows OOD samples to be detected in a \emph{posthoc} manner, \ie it does not require retraining of the base network~\cite{yang2022openood} and can be applied to any pre-trained object detector with a backbone containing \methodname layers (\eg ResNet~\cite{He_2016_CVPR} and RegNetX~\cite{RegNetX}).
    \item We train this MLP on the surrogate task of distinguishing the SAFE vectors of adversarially-perturbed samples and clean ID training samples. This avoids the necessity of access to real outlier training data \cite{Mahalanobis, FSSD, MCD, liu2020energy, OutlierExposure, fixed_MD, sim_seg, OODL} or a complex generative process to synthesise such data \cite{GAN-Synthesis, viaGeneration, uniform_pred, sabokrou2018adversarially}. 
\end{enumerate}

\noindent \methodname achieves new state-of-the-art results across multiple established benchmarks. We release a publicly available code repository to replicate our experiments at: \url{https://github.com/SamWilso/SAFE\_Official}

\section{Related Work}
In this section, we identify the core contributions in the related areas to OOD object detection: i) Many methods attempt to calibrate the confidence of the network utilising available or self-generated outlier data. ii) When access to outlier data is unavailable, deep features of the network are monitored for deviations from known values. iii) Under some regularisation constraints, the feature space of a deep network can be tuned to be \emph{distance-aware}, improving OOD detection performance. iv) Whilst work on OOD object detection is scarce, recent works have been proposed for adjacent tasks (\eg open-set, performance monitoring) in object detection.

\textbf{Outlier-based OOD Detection} A common approach to OOD detection is to calibrate the model confidence by tuning the weights or hyperparameter on an auxiliary validation dataset~\cite{Mahalanobis, FSSD, MCD, liu2020energy, OutlierExposure, fixed_MD, sim_seg, OODL}. These OOD-specific characteristics can be extrapolated from an available set of real outlier data constructed from either the testing OOD set~\cite{FSSD, Mahalanobis, liu2020energy} or an entirely separate dataset~\cite{fixed_MD, sim_seg, OutlierExposure, OODL}. While these methods often present impressive performance, the use of these outlier sets is inherently problematic: if the real outlier set does not accurately represent the OOD samples encountered at test time, substantial drops in performance are observed~\cite{sun2021react}. 

To overcome this, many prior works synthesise outliers as a proxy for OOD samples, training a network to distinguish between ID samples and the synthetic outliers~\cite{GANs, GAN-Synthesis, rejectclassifiers, viaGeneration, uniform_pred, sabokrou2018adversarially, vos}. Early works on outlier synthesis focused on Generative Adversarial Networks (GANs)~\cite{GANs}, training a model that generated low ID density samples in the image space for calibrating confidence measures~\cite{GAN-Synthesis}, training a reject class~\cite{rejectclassifiers, viaGeneration} or encouraging uniform predictions on OOD samples~\cite{uniform_pred}. Scaling input-based generative models becomes complex as the fidelity of images increases, and thus feature-based generative methods have been proposed for OOD object detection~\cite{vos}. Synthetic outliers have also been created by adding input-level perturbations to the known ID dataset via adversarial generation~\cite{ODIN, General-ODIN, generalOOD}, pixel-level mutation~\cite{likelihoodratio} or permutation~\cite{NMD} and additive noise~\cite{mahmood2021multiscale, CSI}. We adopt a similar approach and generate synthetic outliers via adversarial perturbations with the goal of training an auxiliary MLP to distinguish between ID samples and adversarial samples.

\textbf{Feature-based OOD Detection} Many methods avoid generating sufficiently realistic outliers by directly monitoring the outputs~\cite{hendrycks17baseline, hendrycks2019anomalyseg, pmlr-v70-guo17a, Gal2016Dropout, Zaeemzadeh_2021_CVPR, ViM, GradNorm} or features~\cite{pmlr-v119-sastry20a, HDFF, NMD, DUQ, OODL, KNNOOD} of the DNN. These methods are generally more computationally efficient in contrast to alternative OOD detectors, but often rely on fundamental assumptions about the characteristics of the feature space \eg separability of classes in feature space~\cite{PAC-LEARNABLE-OOD, HDFF, Zaeemzadeh_2021_CVPR, DUQ, KNNOOD}. 
Following these findings, recent works have proposed methods that enable the assessment of layer-wise performance, subsequently demonstrating that not all layers are equally effective at detecting OOD data~\cite{HDFF, NMD, pmlr-v119-sastry20a, mahmood2021multiscale} and that some layers exhibit abnormal behaviour when presented with OOD data~\cite{sun2021react}. All these works are applied exclusively to image classification. In this work, we extend the usage of feature-based OOD detectors to object detection by only leveraging the backbone features that are the most \emph{sensitive} to OOD data.

\textbf{Feature Space Regularisation}
The beneficial properties of \textit{sensitivity} and \textit{smoothness} have recently been highlighted in the context of OOD detection for classification~\cite{SNGP, DDU, DUQ, ResnetInputDistance}. \emph{Sensitivity} ensures that differences in the input space (\ie pixels) result in sufficiently different representations in the feature space, preventing the feature collapse problem~\cite{DUQ}. \emph{Smoothness} prevents the feature mapping from being \emph{too} sensitive, thus avoiding reduced generalisation and robustness~\cite{DUQ}. Both properties constitute the lower and upper bounds of a bi-Lipschitz constraint (Eq.~(\ref{eq:bi-lip})) and can be enforced during training, \eg by training a network with residual connections~\cite{He_2016_CVPR} with spectral normalisation~\cite{miyato2018spectral}, as applied by~\cite{DDU, SNGP}. However, recent work~\cite[Sec.~C.4]{DDU} has shown that residual connections constitute an inductive bias towards sensitivity, even \emph{without} spectral normalisation. We build upon this insight and expand it to the task of OOD for object detection.

\begin{figure*}[t]
    \centering
    \includegraphics[width=\textwidth]{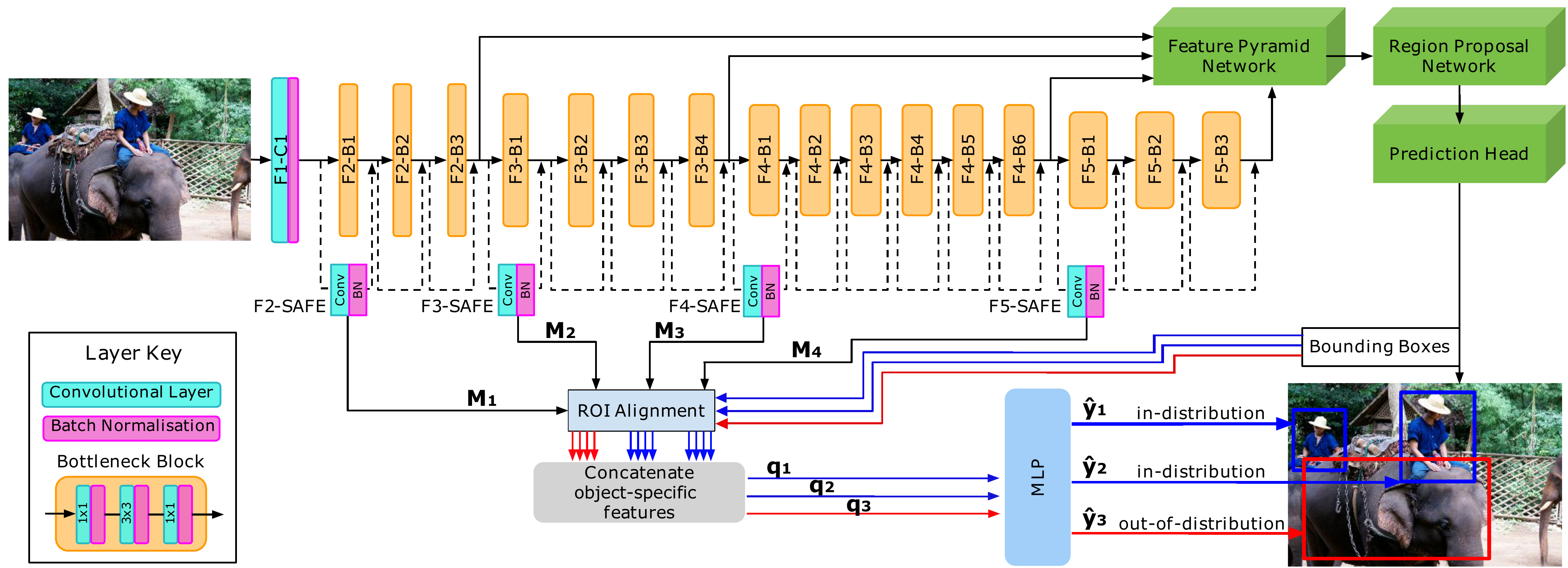}
    \vspace{0.1cm}
    \caption{Architecture diagram of our proposed \methodname OOD detector with an example ResNet-50~\cite{He_2016_CVPR}. We extract object-level feature maps $\{\vM_1,\vM_2,\vM_3,\vM_4\}$ from the layers that are most sensitive to OOD data. Next, object descriptors $\{\vq_1,\vq_2,\vq_3\}$ are formed by applying region of interest pooling to the feature maps and concatenating the resultant vectors layer-wise. Each object descriptor is passed through the MLP, producing producing a corresponding OOD score for each object $\{\hat{y}_1,\hat{y}_2,\hat{y}_3\}$ which distinguishes detections with OOD samples (\textcolor{Red}{red}) from detections with ID samples (\textcolor{Blue}{blue}). 
    }
    \label{fig:process}
\end{figure*}

\textbf{Reliable Object Detection} Applications of OOD detection methods to object detection is a new field; however, there are existing works in related domains. Akin to OOD detection, open-set error detection~\cite{9093355} commonly relies on the outputs of the final layer of the object detector network~\cite{GMM-Det, joseph2021open, miller2018dropout}. In a similar vein, recent works have sought to explain failures in deep object detectors by analysing the relationship between individual architecture components and unique errors~\cite{9813388} and the influence of the datasets on these errors~\cite{Miller_2022_CVPR}. Related works in failure monitoring use auxiliary networks trained on backbone object detector features~\cite{8968525, DBLP:journals/corr/abs-2011-07750}. Few works~\cite{vos, du2022stud, du2022siren} have explicitly addressed the problem of OOD object detection. These methods require explicit retraining of the base network or fine-tuning of the hyperparameters with an auxiliary outlier dataset~\cite{vos, du2022stud, du2022siren}. 
\section{SAFE: Sensitivity-Aware Features}\label{sec:method}
    We propose SAFE, a post-hoc OOD detector for object detection that leverages the \emph{sensitivity} of residual convolutional layers and abnormal batch normalisation activations to identify OOD object detections. In Section~\ref{sec:theory}, we provide the motivation that informs the critical aspects of SAFE, and then in Section~\ref{sec:method:imp}, we introduce the SAFE method in detail.

    \subsection{Motivation} \label{sec:theory}
        Fundamental to \methodname, we identify residual convolution layers followed immediately by batch normalisation are consistently \emph{sensitive} and thus powerful layers for OOD object detection. Whilst it has been shown that not all layers perform equally for OOD detection in image classification ~\cite{HDFF, NMD, pmlr-v119-sastry20a, OODL} or LIDAR-based OOD detection~\cite{LIDAR-OOD}, we are the first to empirically validate this for OOD object detection and the first to investigate the layer characteristics that induce stronger performance. We select our subset of critical layers based on prior work in the image classification setting demonstrating \emph{sensitivity}-preserving properties of residual connections~\cite{DDU} and abnormal activations of batch normalisation layers~\cite{sun2021react}. In the following, we detail the theoretical groundwork underpinning our findings.
        
        \textbf{Sensitivity of Residual Connections} We consider a frozen pre-trained base network $f$ that functions as a feature extractor, mapping samples from the input space $\mathbb{X}$ to the hidden feature space $f: \mathbb{X} \rightarrow \mathbb{H}$. To detect OOD samples, the DNN's feature space needs to be well-regularised~\cite{SNGP}, according to the \textit{bi-Lipschitz constraint}:
        \begin{equation} \label{eq:bi-lip}
            L_1 \cdot \lVert x - x^{*} \rVert_{I} \leq \lVert f(x) - f(x^{*}) \rVert_{F} \leq L_2 \cdot \lVert x - x^{*} \rVert_{I}
        \end{equation}
        where $x$ and $x^{*}$ are two input samples, $\lVert\cdot\rVert_{I}$ and $\lVert\cdot\rVert_{F}$ denote distance metrics in the input and feature space respectively, and $L_1$ and $L_2$ are the lower and upper Lipschitz constants~\cite{SNGP}. The lower bound, \textit{sensitivity}, ensures that distances in the input space are sufficiently preserved in the hidden space, and the upper bound, \textit{smoothness}, limits the sensitivity of the hidden space to input variations, ensuring that distances in the hidden space have a meaningful correspondence to distances in the input space. Encouraging sensitivity and smoothness is commonly accomplished by applying spectral normalisation~\cite{miyato2018spectral} to the weight matrices of a DNN with residual connections~\cite{SNGP, DDU, ResnetInputDistance}. In posthoc OOD detection, where $f$ is pretrained, we cannot guarantee that a network is pretrained with spectral normalisation, hence not fulfilling the \textit{smoothness} constraint. However,~\cite[Sec C.4]{DDU} has shown that a network solely trained \emph{with residual connections} and no \textit{smoothness} constraint is still sufficiently \textit{sensitive} to changes in the input.
        
        \textbf{Mismatched BatchNorm Statistics} Batch Normalisation~\cite{batchnorm} (BatchNorm) is a commonly used normalisation technique to help training deep networks. BatchNorm assists the network in learning the designated task on ID data by normalising a given input $z$ with respect to the expected value $\mathbb{E}_{in}[\cdot]$ and variance $\mathbb{V}_{in}[\cdot]$ calibrated over the ID data:
        \begin{equation}
            \operatorname{BatchNorm}(z;\gamma, \beta, \epsilon) = \dfrac{z-\mathbb{E}_{in}[z]}{\sqrt{\mathbb{V}_{in}[z] + \epsilon}} \cdot \gamma + \beta.
        \end{equation}
        Recent work~\cite{sun2021react} empirically observed that BatchNorm statistics calibrated on the ID set and directly applied to the OOD set trigger abnormally high activations due to a mismatch of the true parameters between datasets $\mathbb{E}_{in}, \mathbb{V}_{in} \neq \mathbb{E}_{out}, \mathbb{V}_{out}$. Propagation of these abnormal activations throughout the network results in abnormally high logits for an erroneous prediction, resulting in overconfidence of the network on OOD samples. We propose a deep feature-based approach to leverage this characteristic; training an auxiliary network to monitor feature activations from these layers and flagging a sample as OOD when an abnormal activation is detected.

    \subsection{Method} \label{sec:method:imp}
        Given the observations that residual connections enable sensitivity to input changes~\cite{DDU} and that BatchNorm layers trigger abnormal activations on OOD data~\cite{sun2021react}, we thus hypothesise that residual connections combined with BatchNorm regularisation provide a clear signal for OOD detection. Connections of this variety are not uncommon, with four of these layers in the standard ResNet-50~\cite{He_2016_CVPR} and RegNetX4.0~\cite{RegNetX} backbone architectures. We now detail the pipeline for \methodname to leverage these critical layers (see Figure~\ref{fig:process}) and later confirm our hypothesis empirically in Section~\ref{sec:layerimp}.
        
        \textbf{Preliminaries} 
        We consider a pretrained, frozen DNN object detector $f$, which given an image $x$ will produce a set of $D$ object predictions. Each detection $d \in \{1, ...,D\}$ has a classification label $c_d$ and bounding box $b_d \in \mathbb{R}^4$. During deployment, we wish to classify each object prediction as being generated from ID or OOD data.

        \textbf{Object-specific SAFE Extraction} To distinguish between ID and OOD object predictions, we extract object-specific features from our identified sensitivity-aware layers. In the detector, there are $L$ \methodname layers, \ie residual connection layers with BatchNorm regularisation, that output a set of $L$ feature maps $\{\vM_1,...,\vM_L\}$. Figure~\ref{fig:process} shows an example of an object detector with a ResNet-50 backbone, which contains $L=4$ \methodname layers. To extract object-specific features, the proposed bounding boxes $\{b_1, \dots, b_D\}$ are used to take cropped regions of each feature map $\vM_l$. These object-specific feature maps $\vO_{l, d}$ are then reduced to a vector representation $\vp_{l, d}$ for concatenation via a bilinear interpolation operation along the spatial axis. Finally, the pooled feature vectors $\vp_{l, d}$ are concatenated layer-wise to form a single object-specific vector $\vq_d$ with a length equal to the sum of the number of channels $c$ for each layer: $|\vq_d|=\sum_l{c_l}$.
        
        \textbf{Feature Monitoring MLP} We instantiate an auxiliary feature monitoring MLP, $f_\beta$, to classify detections as ID or OOD. Given an object-specific \methodname vector, the MLP outputs an OOD score $\hat{y}_d=f_\beta(\vq_d)$ in the range of $\hat{y} \in [0, 1]$. These scores can be used in practice to make decisions based on the application and corresponding risk profile of the downstream task, \eg detection scores can be compared to a predefined threshold to classify objects as ID or OOD and achieve a minimum true positive rate.
        
        \textbf{Surrogate Training with Synthetic Outliers} Since OOD samples are inaccessible prior to deployment, the feature monitoring MLP is trained on the \emph{surrogate} task of discriminating between objects within clean ID images and the same objects within adversarially-perturbed ID images. For each image in the training set $x \in \mathbb{X}$, we generate an outlier counterpart of the same image through an adversarial perturbation $x^o=g(x)$. In practice, we utilise the simple Fast Gradient Sign Method~\cite{FGSM} (FGSM) adversarial attack that produces a perturbed image $x^o$ by adding noise to the original image $x$. Given the model parameters $\theta$, the noise for FGSM is computed based on the sign of the gradient $\operatorname{sign}(\nabla_{x})$ with respect to the cost function $J(\theta,x,y))$ and then scaled with by magnitude multiplier $\epsilon$:
        \begin{equation}
            x^o= x + \epsilon \cdot \operatorname{sign}\big(\nabla_{x}J(\theta,x,y)\big).
        \end{equation}
        Next, object-specific feature vectors are extracted from the SAFE layers for both the clean and perturbed images using the bounding boxes $b_d$ predicted \emph{on the clean image}. Finally, the object-specific feature vectors are used to train the auxiliary MLP with clean features corresponding to ID detections and perturbed features corresponding to surrogate OOD detections. We ablate the parameters of our adversarial-perturbation in Section~\ref{sec:magnitude}.%

\section{Experiments}
We conduct a series of experiments to demonstrate the efficacy of our proposed \methodname OOD detector. We first describe our experimental setup in Section~\ref{sec:experimentalsetup} and detail the implementation of \methodname in Section~\ref{sec:implementation}. We then compare our method to the state-of-the-art on the challenging task of OOD object detection in Section~\ref{sec:results}. Finally, we demonstrate the unique effectiveness of our identified critical layers in Section~\ref{sec:layerimp}, and we ablate the sensitivity of the auxiliary MLP to the transformation magnitude in Section~\ref{sec:magnitude}. Additional comparisons of \methodname to the state-of-the-art with the transformer-based Deformable DETR object detector~\cite{zhu2021deformable} are provided in the Supplementary Material.

\subsection{Experimental Setup}
\label{sec:experimentalsetup}
We follow the evaluation protocol defined by~\cite{vos} with the accompanying benchmark repository\footnote{https://github.com/deeplearning-wisc/vos}.

\textbf{Datasets} We use the predefined ID/OOD splits for the object detection task defined in~\cite{vos}. The two ID datasets are constructed from the popular PASCAL-VOC~\cite{PASCAL-VOC} and Berkley DeepDrive-100K~\cite{BDD100K} (BDD100K) datasets. For the OOD datasets, subset versions of the MS-COCO~\cite{MS-COCO} and OpenImages~\cite{OpenImages} datasets are provided where classes that appear in the custom ID datasets are removed. %

\textbf{Evaluation Metrics} We consider the standard AUROC and FPR95 metrics defined in~\cite{vos} extensively used across the image classification literature~\cite{Zaeemzadeh_2021_CVPR, ODIN, HDFF, pmlr-v119-sastry20a}. \textbf{AUROC:} The Area Under The Receiver Operating Characteristic curve (AUROC) is defined by the area under the ROC curve with true positive rate (TPR) on the y-axis and false positive rate (FPR) on the x-axis; higher is better. An AUROC score of 50\% indicates a method that is as effective as random guessing. \textbf{FPR95} reports the false positive rate when the true positive rate is at 95\%; lower is better. For real-world deployment, a binary classifier based on a threshold of the confidence scores determines if a detection is ID or OOD; under these conditions, FPR95 provides better insight into how an OOD detector will perform.  \textbf{AP:} Our \methodname OOD detector is a \emph{posthoc} addition to a pre-trained network and does not affect the on-task performance of the base model under the average precision (AP) metric, as such, we do not report AP as in~\cite{vos}. 

Importantly, under the benchmark setting by~\cite{vos}, the AUROC and FPR95 metrics are computed \emph{after} low-confidence objects are suppressed according to a confidence threshold determined by~\cite{harakeh2021estimating}. Our comparisons in Table~\ref{tab:eval} implement this suppression for fair comparisons while our \emph{qualitative} visuals in Figure~\ref{fig:supp:qual} visualise some low-confidence detections.

\textbf{Random Seeds} As there is inherent randomness in the initialisation of the auxiliary MLP, we report the mean $\mu$ and standard deviation $\sigma$ of each metric over five seeds; the default benchmark seed~\cite{vos} (0) and four randomly generated seeds in the range of $[1, 10^5]$ for replicability, in the format of $\mu \pm \sigma$. 

\textbf{Baselines} We compare against the following state-of-the-art methods: MSP~\cite{hendrycks17baseline}, ODIN~\cite{ODIN}, Mahalanobis Distance~\cite{Mahalanobis}, Energy Score~\cite{liu2020energy}, Gram Matrices~\cite{pmlr-v119-sastry20a}, ViM~\cite{ViM}, KNN~\cite{KNNOOD}, Generalized ODIN~\cite{General-ODIN}, CSI~\cite{CSI}, GAN-Synthesis~\cite{GAN-Synthesis} and Virtual Outlier Synthesis (VOS)~\cite{vos}. Performance metrics for ViM and KNN are reported from implementations based on public code. %
Performance metrics for all other methods are reported from~\cite{vos}. 

\subsection{Implementation}
\label{sec:implementation}
\textbf{Base Network Architecture} Following the evaluation protocol defined in~\cite{vos}, we implement the Faster-RCNN~\cite{Faster-RCNN} detector with either a ResNet-50~\cite{He_2016_CVPR} or RegNetX4.0~\cite{RegNetX} backbone using the Detectron2 library~\cite{wu2019detectron2}. All compared methods, excluding VOS~\cite{vos}, are evaluated exclusively using the ResNet-50 backbone consistent with~\cite{vos}; VOS and \methodname are compared on both the ResNet-50 and RegNetX4.0 backbones. Of the compared methods, Generalized ODIN~\cite{General-ODIN}, CSI~\cite{CSI}, GAN-Synthesis~\cite{GAN-Synthesis} and VOS~\cite{vos} all require the base object detector to be retrained with a custom loss objective, we identify these methods with a checkmark \checkmark~in Table~\ref{tab:eval}. For a fair comparison, we report the results of \methodname and VOS~\cite{vos} using both the ResNet-50 and RegnetX4.0 backbones to ensure that the differing on-task performance, which has been shown to affect open-set recognition performance~\cite{vaze2022openset}, does not bias the results. 

\textbf{Feature Extraction} During feature extraction, hooks are applied to the output of the critical residual + BatchNorm layer combinations within the ResNet-50 and RegNetX4.0 backbones of the Faster-RCNN model. Object-specific features $\vp_{l,d}$ are retrieved using the ROIAlign~\cite{ROIAlign} operation with the predicted bounding boxes $b$. Appropriate spatial scaling factors in ROIAlign are set so that features are pooled to a channels length $c_l$ vector per layer $l$.
\begin{table*}[ht]
    \setlength{\tabcolsep}{2.5pt} %
    \centering
    \small
    \begin{tabular}{l|c|cccc|cccc}
        &  & \multicolumn{4}{c|}{\textbf{ID: PASCAL-VOC}} & \multicolumn{4}{c}{\textbf{ID: Berkley DeepDrive-100K}} \\
        \multicolumn{1}{l|}{\textbf{Method}} & \textbf{Retrain?} &  \multicolumn{2}{c}{\textbf{OpenImages}} & \multicolumn{2}{c|}{\textbf{MS-COCO}} & \multicolumn{2}{c}{\textbf{OpenImages}} & \multicolumn{2}{c}{\textbf{MS-COCO}} \\ \cline{3-10}
        & & AUROC$\uparrow$ & FPR95$\downarrow$ & AUROC$\uparrow$ & FPR95$\downarrow$ & AUROC$\uparrow$ & FPR95$\downarrow$ & AUROC$\uparrow$ & FPR95$\downarrow$ \\ \hline %
         MSP \cite{hendrycks17baseline} & & 81.91  & 73.13 & 83.45 & 70.99  & 77.38 & 79.04 & 75.87 &  80.94 \\
         ODIN \cite{ODIN} & & 82.59 & 63.14 & 82.20 & 59.82 & 76.61 & 58.92 & 74.44 &  62.85 \\
         Mahalanobis \cite{Mahalanobis} & & 57.42 & 96.27 & 59.25 &  96.46 & 86.88 & 60.16 & 84.92 &  57.66\\
         Energy Score \cite{liu2020energy} & & 82.98 & 58.69 & 83.69 &  56.89 & 79.60 & 54.97 & 77.48 &  60.06\\
         Gram Matrices \cite{pmlr-v119-sastry20a} & & 77.62 & 67.42 & 79.88 & 62.75 & 59.38 & 77.55 & 74.93 & 60.93 \\
         ViM \cite{ViM} & & 68.73 & 88.40 & 71.94 & 83.47 & 86.49 & 53.80 & 87.17 & 54.58 \\
         KNN \cite{KNNOOD} & & 85.08 & 55.73 & 86.07 & 54.50 & 88.37 & 44.50 & 87.45 & 47.28 \\
         Generalized ODIN \cite{General-ODIN} & \checkmark & 79.23 & 70.28 &  83.12 & 59.57 & 87.18  & 50.17 & 85.22 & 57.27\\
         CSI \cite{CSI} & \checkmark & 82.95 & 57.41 & 81.83 &  59.91 & 87.99 & 37.06 & 84.09 &  47.10\\
         GAN-Synthesis \cite{GAN-Synthesis}& \checkmark & 82.67 & 59.97 & 83.67 & 60.93 & 81.25 & 50.61 & 78.82 &  57.03 \\
         VOS-ResNet50 \cite{vos} & \checkmark & 85.23$\pm$0.6 & 51.33$\pm$1.6 & \second{88.70}$\pm$1.2 & 47.53$\pm$2.9 & 88.52$\pm$1.3 & 35.54$\pm$1.7 & 86.87$\pm$2.1 &  44.27$\pm$2.0 \\
         VOS-RegNetX4.0 \cite{vos} & \checkmark & 87.59$\pm$0.2 & 48.33$\pm$1.6 & \best{89.00}$\pm$0.4 & 47.77$\pm$1.1 & 92.13$\pm$0.5 & 27.24$\pm$1.3 & \second{89.08}$\pm$0.6 & 36.61$\pm$0.9 \\ 
        \hline 
        \textbf{\methodname-ResNet50 (ours)} & & \second{92.28}$\pm$1.0 & \second{20.06}$\pm$2.3 & 80.30$\pm$2.4 & \second{47.40}$\pm$3.8 & \second{94.64}$\pm$0.3 & \second{16.04}$\pm$0.5 & 88.96$\pm$0.6 & \second{32.56}$\pm$0.8 \\
        
        \textbf{\methodname-RegNetX4.0 (ours)} & & \best{94.38}$\pm$0.2 & \best{17.69}$\pm$1.0 & 87.03$\pm$0.5 & \best{36.32}$\pm$1.1 & \best{95.97}$\pm$0.1 & \best{13.98}$\pm$0.3 & \best{93.91}$\pm$0.1 & \best{21.69}$\pm$0.5 \\

    \end{tabular}\vspace*{0.15cm}%
    \caption{OOD detection results comparing \methodname to state-of-the-art OOD detectors. Comparison metrics are FPR95 and AUROC, directional arrows indicate if higher ($\uparrow$) or lower ($\downarrow$) values indicate better performance. \best{Best} results are shown in \best{red and bold}, \second{second} best results are shown in \second{orange}. Methods that require retraining are indicated with a checkmark \checkmark. Mean and standard deviation over 5 seeds is shown for \methodname. We observe that \methodname provides strong performance across almost all benchmarks and metrics, achieving the highest performance across 7 out of 8 of the benchmark permutations. Notably, we observe substantial reductions in FPR95, particularly when OpenImages is the OOD set, with a greater than 30\% reduction for both backbones under the PASCAL-VOC setting.
    }
  \label{tab:eval}%
\end{table*}%

\textbf{MLP Architecture} Following previous works on auxiliary network feature monitoring~\cite{NMD}, the auxiliary MLP is constructed as a 3-layer fully connected MLP with a single output neuron fed into a Sigmoid activation with a dropout connection before the final layer. The size for each fully connected layer is progressively halved with each consecutive layer. The MLP, initialised with Xavier initialisation~\cite{Xavier-init}, is trained for 5 epochs using binary cross entropy loss optimised by SGD with a learning rate of $10^{-3}$, momentum of 0.9, dropout rate of 50\% and batch size of 32 images\footnote{The size of each individual batch for the MLP is determined by the number of predicted boxes within the 32 images.}.

\textbf{Transform Implementation} We implement FGSM~\cite{FGSM}, parameterised by a scalar magnitude multiplier $\epsilon$, as our adversarial-perturbation for the surrogate MLP training task. During comparisons in Section~\ref{sec:results}, we set $\epsilon=8$ when ResNet-50 is the backbone and $\epsilon=1$ for RegNetX4.0. We ablate the sensitivity to $\epsilon$ on ResNet-50 in Section~\ref{sec:magnitude}. %

\subsection{Results and Discussion} \label{sec:results}
    Table~\ref{tab:eval} compares the performance of our \methodname detector to the current state-of-the-art in OOD object detection. \methodname sets a new state-of-the-art across 7 out of the 8 benchmark permutations. We observe substantial reductions to the FPR95 metric, with the OpenImages as OOD setting improving by more than 30\% when PASCAL-VOC is ID and 20\% when BDD100K is the ID set, with the most significant differences when comparing directly between ResNet-50 models. These observations are further substantiated when considering \methodname in contrast to other posthoc OOD detectors with substantial performance improvements across the majority of metrics, exemplified by improvements of $\sim$35\% in FPR95 under the OpenImages setting for both datasets. In summary, \methodname, which does not require retraining, outperforms OOD detectors \emph{that do require retraining}, and significantly outperforms other posthoc OOD detectors.
    
    \textbf{Robustness} We further note that the results from Table~\ref{tab:eval} demonstrate the robustness of \methodname to varying model architectures (\ie ResNet-50 and RegNetX4.0), given that the target models contain the specified critical layers as discussed in Section~\ref{sec:theory}. We reiterate that \methodname does not require a specified training regime and thus both networks are trained without a specialised loss. Directly comparing between \methodname and VOS~\cite{vos} on the same ResNet-50 backbone, we observe that \methodname outperforms VOS across all metrics under the BDD100K setting and the majority of metrics when PASCAL-VOC is ID. Under the architectural shift towards the RegNetX4.0 backbone, we observe that \methodname still retains high performance, outperforming VOS under the majority of metrics under the PASCAL-VOC setting and providing higher AUROC and FPR95 results for both OOD sets under the BDD100K setting. %

    \begin{figure*}[ht]
    \centering
    \includegraphics[width=\textwidth]{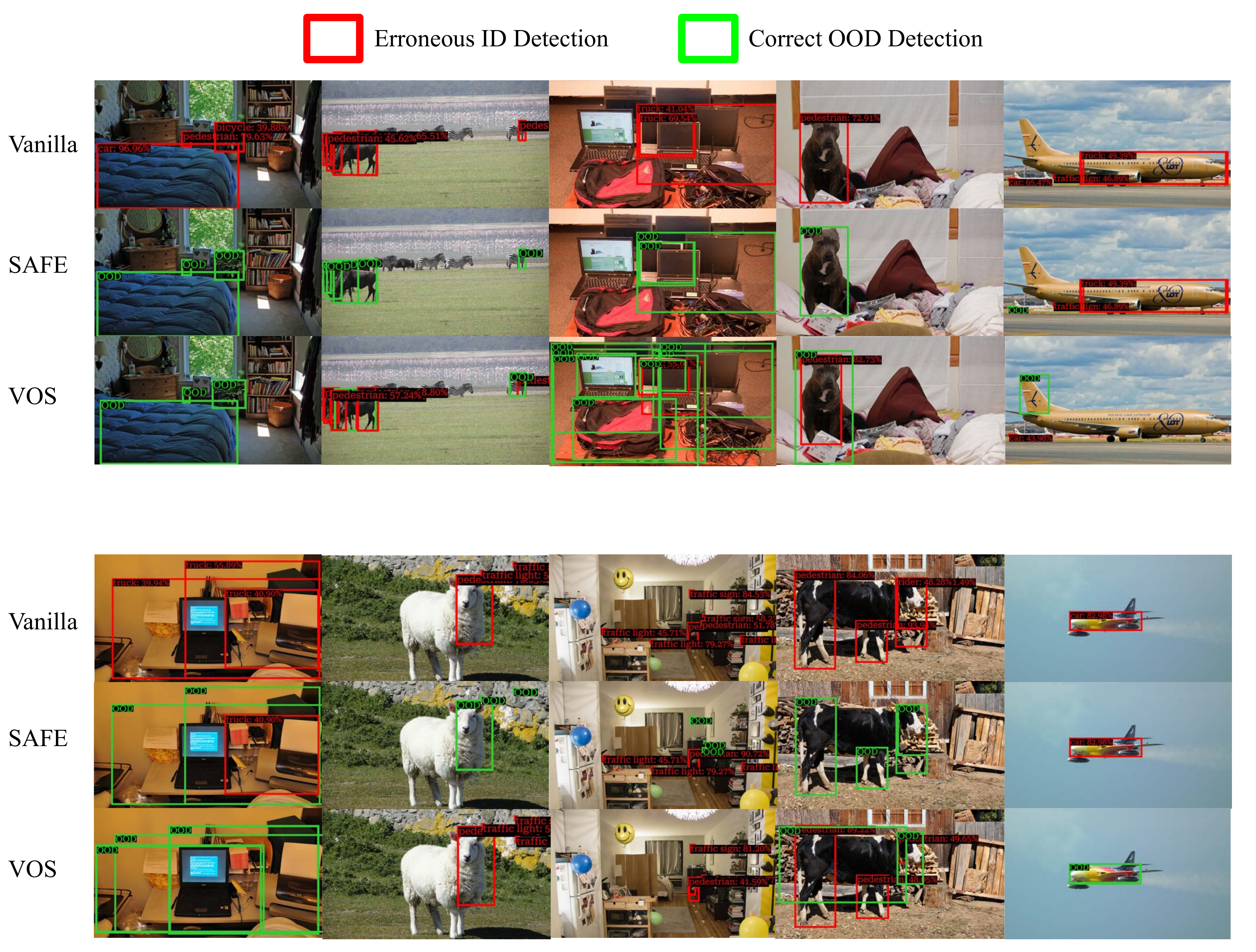}
    \caption{Qualitative visualisation of object detections from the ResNet-50 Faster-RCNN on samples from the MS-COCO OOD dataset. BDD100K is the ID dataset. Binarisation of OOD detection scores are achieved by thresholding the OOD scores using the same threshold used to compute the FPR95 metrics. \textcolor{ForestGreen}{Green} bounding boxes signify detections that were \emph{correctly} flagged as OOD and \textcolor{red}{red} detections are \emph{incorrectly} considered ID. \textbf{Top}: Vanilla predictions from the object detector. \textbf{Middle}: OOD detections by \methodname. \textbf{Bottom}: OOD detections by VOS~\cite{vos}. 
    }
    \vspace*{-0.4cm}
    \label{fig:supp:qual}
\end{figure*}

    \textbf{Qualitative Results} Figure~\ref{fig:supp:qual} visualises the object predictions of the base network (Top) and subsequent OOD detections from \methodname (Middle) or VOS~\cite{vos} (Bottom) on a set of MS-COCO test images when the ID dataset is BDD100K. We observe that \methodname successfully identifies many of the OOD objects within the scenes, reducing the impact of these erroneous predictions during deployment. 
    
    Consistent with the quantitative results from Table~\ref{tab:eval}, we observe that \methodname is more reliable at detecting OOD samples than VOS~\cite{vos}. In particular, we observe that in some instances VOS generates additional erroneous predictions (Figure~\ref{fig:supp:qual}, Columns 2 \& 4), flagging only a subset of these instances as OOD. In contrast, \methodname correctly detects all of the object instances predicted by the vanilla network as OOD in these images.
    
    We note that \methodname is susceptible to some failures where an object may have similar features to an ID class. The right-most column of Figure~\ref{fig:supp:qual} provides two examples of this where the base network predicts vehicle labels (truck/car) onto an airplane which \methodname does not detect as erroneous. 

\subsection{Layer Importance} \label{sec:layerimp}
    Fundamental to the theory of our proposed \methodname detector (Section~\ref{sec:theory}) is the importance of residual and BatchNorm layers. Critically, we leverage theoretical and empirical foundations for residual connections enabling \textit{sensitivity} of the network~\cite{DDU} and BatchNorm layers triggering abnormal activations on OOD data~\cite{sun2021react} to address OOD object detection. 
    
\begin{figure}[t]
    \centering
    \includegraphics[width=\columnwidth]{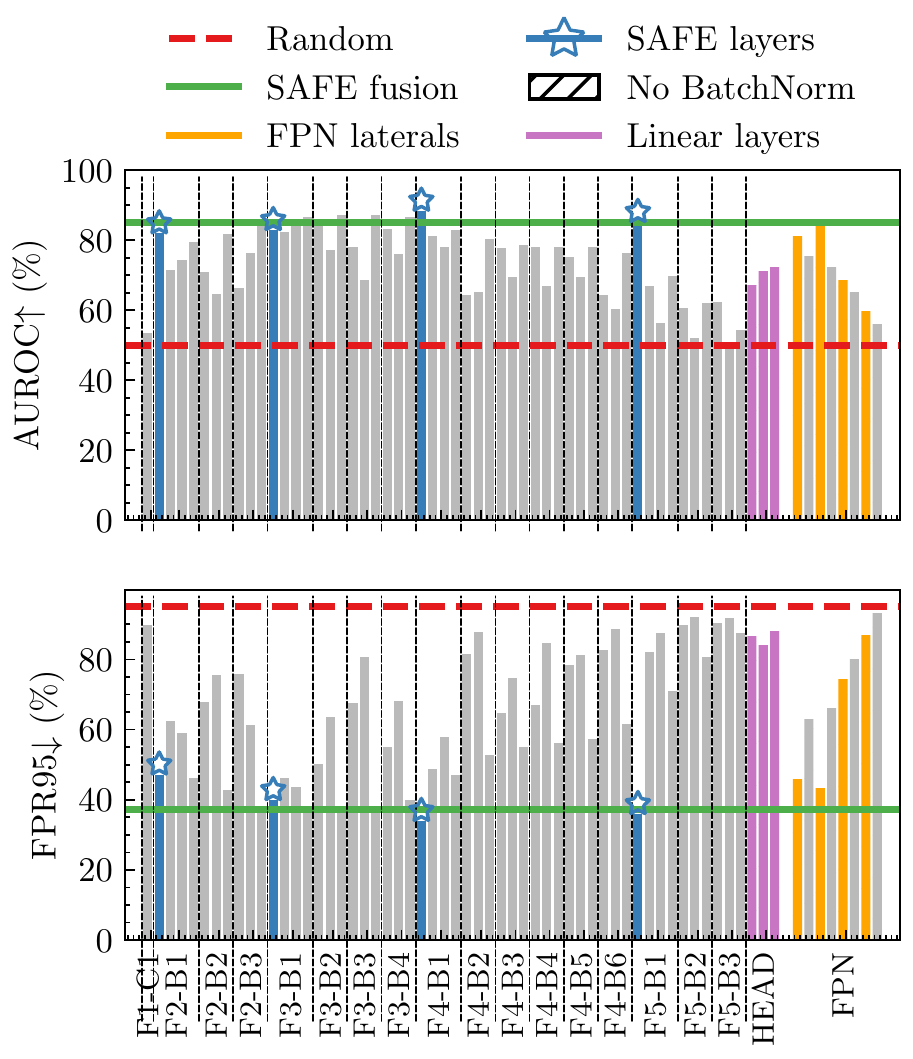}
    \caption{OOD detection performance of individual Conv2d layers in the standard ResNet-50 backbone (see Figure~\ref{fig:process}) when PASCAL-VOC is the ID set. \textbf{Top}: Comparison metric is AUROC, higher is better. \textbf{Bottom}: Comparison metric is FPR95, lower is better. Results are reported as averages over both OOD datasets. Layers in \textcolor{Blue}{blue} with a star are the identified critical layers for \methodname. Striped layers belong to the Feature Pyramid Network (FPN) and are the only Conv2d layers that \emph{do not have BatchNorm applied immediately after}. \textcolor{purple}{Purple} layers are the fully-connected layers of the Faster-RCNN object detector head. The SAFE critical layers consistently provide among the highest performance across all layers within the ResNet-50 backbone.
    }

    \label{fig:layer}
\end{figure}

    We expand upon these foundations by considering residual convolution + BatchNorm combinations which we expect to leverage the characteristics of both; triggering abnormal activations on OOD inputs which the auxiliary MLP consequently detects. Therefore, we expect that layers that do not satisfy \emph{both} the residual convolution and BatchNorm combinations will not perform as effectively as those layers that do. We empirically verify this hypothesis by ablating the performance of individual layers (Figure~\ref{fig:layer}) and sampling random layer subsets with increasing size (Table~\ref{tab:layer}). We provide expanded versions of these ablations with an additive noise input perturbation in the Supplementary Material.
    
    \textbf{Individual Layer Performance} Figure~\ref{fig:layer} ablates the performance of individual Conv2d layers of the ResNet-50 backbone as the average over both OOD datasets under the AUROC (Figure~\ref{fig:layer}, top) and FPR95 (Figure~\ref{fig:layer}, bottom) metrics when PASCAL-VOC is the ID set. Our identified residual convolution + BatchNorm (SAFE) layers are among the highest performing layers in the network. The majority of the other Conv2d layers report low performance, with very few layers having comparable performance to the SAFE critical layers. Residual connections alone are insufficient as there is no consistently high performance at the beginning of ConvBlocks (separated by vertical dashed lines) which take the added residual from the previous ConvBlock as input or the Feature Pyramid Network lateral connections. Similarly, all of the layers outside of the Feature Pyramid Network are followed immediately by a BatchNorm layer, but this alone is insufficient since many of these layers produce poor performance. %
    Figure~\ref{fig:layer} thus provides further empirical evidence, compounding the foundational works in image classification supporting our hypothesis~\cite{sun2021react, ResnetInputDistance, DUQ, DDU, SNGP}, that residual convolution + BatchNorm layer combinations provide powerful OOD detection performance. %

    We observe two further characteristics when inspecting Figure~\ref{fig:layer}: (1) A cluster of relatively high-performing layers between block F3-B1 through to F4-B1 and (2) The highest performing layer in most blocks is the last Conv2d layer. It is not unexpected that there are other high-performing layers other than the SAFE critical layers. Prior works~\cite{HDFF, pmlr-v119-sastry20a, NMD, OODL} in image classification have established that individual layer performance varies dependent on the ID and OOD data distributions. However, with no theoretical foundation for the selection of these layers, \emph{a priori} selection, \ie selection prior to testing, is infeasible. Furthermore, prior works~\cite{HDFF, pmlr-v119-sastry20a, NMD, OODL} suggest that the performance of the non-SAFE layers will vary as the surrogate outlier data distribution shifts; we discuss this with the additive noise outliers in the Supplementary Material. 

    \begin{table}[t]
    \centering
    \small
    \begin{tabular}{|l|cc|cc|}
        \hline & \multicolumn{2}{c|}{\textbf{OpenImages}} & \multicolumn{2}{c|}{\textbf{MS-COCO}} \\
        \textit{Layers} & AUROC$\uparrow$ & FPR95$\downarrow$ & AUROC$\uparrow$ & FPR95$\downarrow$ \\ \hline
        1 & 75.31 & 69.21 & 68.09 & 82.63 \\
        4 & 68.82 & 77.85 & 65.19 & 83.91 \\
        8 & 71.75 & 67.21 & 65.40 & 78.28 \\
        16 & 73.02 & 66.10 & 67.31 & 75.83 \\
        Residuals & \second{91.33} & \second{24.82} & \best{81.87} & \second{48.45} \\
        All (60) & 89.88 & 26.73 & \second{81.30} & 48.57 \\
        \hline
        \textbf{\methodname} & \best{92.28} & \best{20.06} & 80.30 & \best{47.40} \\ \hline
        
    \end{tabular}\vspace*{0.2cm}%
    \caption{Comparison of varied-size layer combinations detecting OOD data when PASCAL-VOC is the ID set. All compared layer subsets \emph{do not contain the identified sensitive layers used in \methodname}. Colour coding and metrics follow those from Table~\ref{tab:eval}. Mean over 5 seeds is shown. We observe that the sensitive layers utilised by \methodname provide disproportionately high performance for OOD detection, outperforming all layer subsets, of which many have access to more than 2x the number of layers as \methodname. Residual layers produce strong performance, comparable to the fusion of all layers, but are still inferior to the fusion of \methodname layers.}
  \label{tab:layer}%
\end{table}%

    \textbf{Layer Subsets} Table~\ref{tab:layer} compares performance of randomly selected subsets of layers \emph{that do not contain any of the identified critical layers} against our \methodname detector with only the four critical layers. As expected from observations of Figure~\ref{fig:layer}, using the SAFE layers significantly outperforms the randomly sampled subsets, even when the subsets contain more layers than SAFE. Subsets of layers perform worse than randomly sampled individual layers, with the performance gap tapering off as the subsets get larger. We attribute this characteristic to the large prevalence of poorly performing layers in the backbone, where the signal from high-performing layers is lowered due to noise from poor-performing layers in the smaller subsets. Using all 60 layers produces better performance than any of the subsets, but still underperforms when compared to our four SAFE layers. %
    
    Consistent with the theory described in Section~\ref{sec:theory}, Table~\ref{tab:layer} demonstrates that the 12 residual connections produce strong performance, performing comparable to the fusion of all Conv2d layers. Whilst the residual connections do provide strong performance, they are outperformed by the \methodname critical layers across both metrics under the OpenImages setting and FPR95 under the MS-COCO setting. 
    
    We note that the size of the auxiliary MLP input scales with the number of layers, and hence feature dimensionality, in the subset. This entails $O(n^2)$ scaling in the weight matrices of the auxiliary MLP, making direct inclusion of large subsets (\eg the 12 residual connections) or all layers computationally expensive. %
    \begin{figure}[t]
    \centering
    \includegraphics[width=0.95\columnwidth]{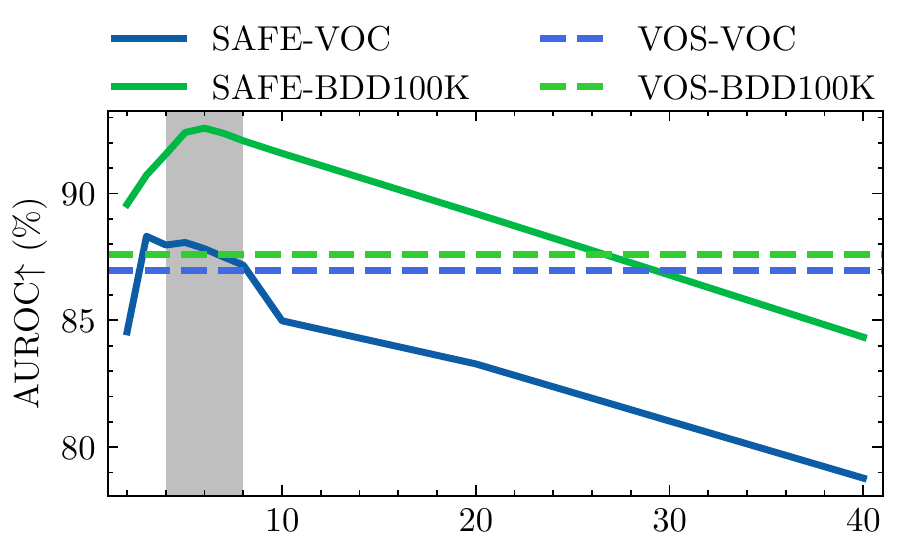}
    \includegraphics[width=0.95\columnwidth]{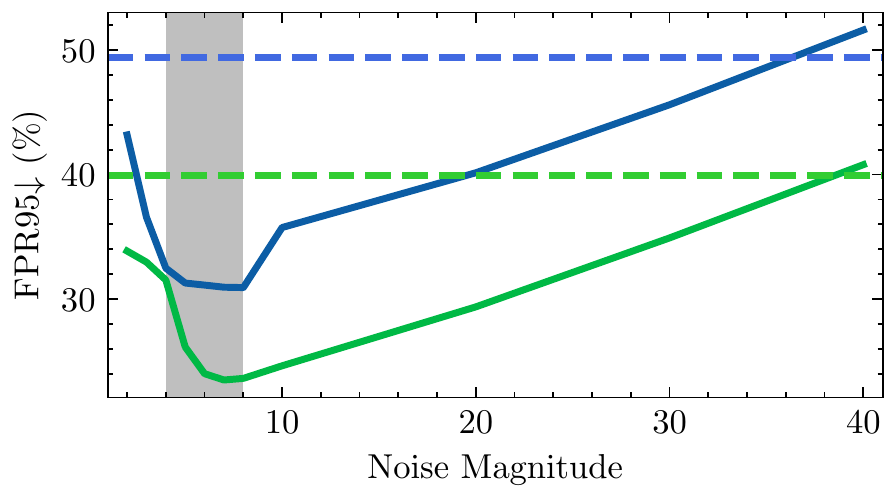}
    \caption{OOD detection performance of \methodname with the ResNet-50 backbone as the gradient sign magnitude $\epsilon$ is varied. \textbf{Top:} Comparison metric is AUROC, higher is better. \textbf{Bottom:} Comparison metric is FPR95, lower is better. Individual lines correspond to the average performance over both OOD sets for the given ID set. Dashed lines correspond to the performance of VOS~\cite{vos} for the respective datasets. A region of consistent high performance for all ID and OOD permutations exists between $\epsilon \in [4, 8]$ (grey region), suggesting that values in and near this range will generalise well to additional datasets.
    }
    \label{fig:weight}
\end{figure}

\subsection{Gradient Magnitude Sensitivity} \label{sec:magnitude}
    Figure \ref{fig:weight} ablates the sensitivity of the auxiliary MLP to varying values of the gradient sign magnitude $\epsilon$ when PASCAL-VOC is the ID set. In general, the performance curves reported match expectations where we observe initial low relative performance due to the MLP being unable to effectively discriminate between the perturbed ID and clean ID features, which improves up to a peak and is followed by a drop in performance as the weighting parameter $\epsilon$ becomes too large, destroying too much of the input content. Critically, we make the observation that a region of high performance exists across all ID, OOD and metric permutations, residing approximately within $\epsilon \in [4, 8]$. The consistently high performance across both ID and OOD dataset permutations suggests that values in this range generalise well to unseen data. 
    
    We further note that Figure~\ref{fig:weight} shows that \methodname generally performs well under a wide range of perturbation magnitudes. Comparing the performance under the FPR95 metric, we observe that only the edge cases of very large values of $\epsilon$ result in worse performance than the previous state-of-the-art. This argument holds particularly true for BBD100K, where a random $\epsilon$ value could be selected in the range of $\epsilon \in [1, 20]$ and \methodname would retain better performance under both AUROC (Figure~\ref{fig:weight}, top) and FPR95 (Figure~\ref{fig:weight}, bottom) than the state-of-the-art, VOS~\cite{vos}.

\section{Conclusion}
In this paper, we propose \methodname, a novel OOD detection framework that leverages the layers in an object detector's backbone that are most \emph{sensitive} to OOD inputs. Unlike previous feature-based OOD object detectors, \methodname leverages the backbone of an object detector network, identifying that the subset of residual convolutions followed by batch normalisation are consistently among the most powerful layers in the network at detecting out-of-distribution samples.

To take advantage of these powerful layers, \methodname trains an auxiliary MLP on the \emph{surrogate} task of distinguishing minimally perturbed adversarial ID samples to clean ID samples using only the features from this subset of layers. We provide a theoretical grounding for the disproportionate power of these layers from image classification literature, expanding upon it to the challenging task of OOD object detection, where we are the first to demonstrate these characteristics. We provide empirical evidence supporting our theory, demonstrating that our identified \methodname layers are among the most powerful layers individually and outperform the fusion of much larger subsets of layers.

\methodname is the first method that considers the \textit{sensitivity} and the impact of individual layers under the setting of OOD object detection. We are optimistic for future work expanding upon our findings through further leveraging our identified sensitive layers, integration of backbone features into OOD object detection, and further theoretical analysis on \emph{sensitivity} and \emph{smoothness} in object detection. We believe that \methodname represents an important step forward in our understanding of OOD object detection and offers a promising avenue for future research.

\ificcvfinal
{\footnotesize
\noindent\textbf{Acknowledgements:} The authors acknowledge continued support from the Queensland University of Technology (QUT) through the Centre for Robotics. TF was partially supported by funding from ARC Laureate Fellowship FL210100156 and Intel Research via grant RV3.290.Fischer. 
}
\fi

{\small
\bibliographystyle{ieee_fullname}
\bibliography{bib}
}

\end{document}